\documentclass[10pt, a4paper]{article}

\usepackage[]{lrec-coling2024}

\usepackage{amsmath}

\name{\parbox{\linewidth}{\centering Chengyuan Liu$^{1,2,\text{\textrm{\dag}}}$\thanks{$^\textrm{\dag}$This work was done when Chengyuan Liu interned at Alibaba.}\hspace{1.5mm}, Yangyang Kang$^{2}$\hspace{1.5mm}, Fubang Zhao$^{2}$\hspace{1.5mm}, Kun Kuang$^{1,\ast}$\thanks{*Corresponding author.}\hspace{1.5mm},\\Zhuoren Jiang$^{3,\ast}$\hspace{1.5mm}, Changlong Sun$^{2}$\hspace{1.5mm}, Fei Wu$^{1,4}$\\}}

\address{liucy1@zju.edu.cn, \{yangyang.kangyy, fubang.zfb\}@alibaba-inc.com, kunkuang@zju.edu.cn\\
         jiangzhuoren@zju.edu.cn, changlong.scl@taobao.com, wufei@zju.edu.cn \\
         $^{1}$College of Computer Science and Technology, Zhejiang University\\
         $^{2}$Institute for Intelligent Computing, Alibaba Group\\
         $^{3}$School of Public Affairs, Zhejiang University\\
         $^{4}$Shanghai Institute for Advanced Study of Zhejiang University}

\usepackage{booktabs}
\usepackage{multirow}
\usepackage{array}
\usepackage{graphicx}
\usepackage{algorithm}
\usepackage{algorithmic}
\usepackage{caption}
\usepackage{subfigure}
\usepackage{amssymb}

\title{Evolving Knowledge Distillation with Large Language Models and Active Learning}

\abstract{
Large language models (LLMs) have demonstrated remarkable capabilities across various NLP tasks. However, their computational costs are prohibitively high. To address this issue, previous research has attempted to distill the knowledge of LLMs into smaller models by generating annotated data. %Nonetheless, these works have mainly focused on text generation and labeling, without deeply exploring knowledge of LLMs for the target task. 
Nonetheless, these works have mainly focused on the direct use of LLMs for text generation and labeling, without fully exploring their potential to comprehend the target task and acquire valuable knowledge.
In this paper, we propose \textsc{EvoKD}: Evolving Knowledge Distillation, which leverages the concept of active learning to interactively enhance the process of data generation using large language models, simultaneously improving the task capabilities of small domain model (student model).
Different from previous work, we actively analyze the student model's weaknesses, and then synthesize labeled samples based on the analysis. In addition, we provide iterative feedback to the LLMs regarding the student model's performance to continuously construct diversified and challenging samples.
Experiments and analysis on different NLP tasks, namely, text classification and named entity recognition show the effectiveness of \textsc{EvoKD}.  \\ \newline  }
\begin{document}

\maketitleabstract

\section{Introduction}\label{sec:intro}
% new architecture 我感觉在intro的部分说明白，先用CoAI那边文章引出black-box kd，用zerogen那篇文章引出blackbox kd就是data generation，然后引出后面的sungen等其他baseline
% 1
% introduce kd, hinton paper: Distilling the Knowledge in a Neural Network
% previous kd still need annotated data, with llm, we dont need human-annotation
% coai black-box kd, white...

% compare with white..., black-box kd advantage: free model structure, powerful black-box llm, lower cost...

% 2
% zerogen first introduce (black-box kd) distill knowledge via data generation
% sungen, balabala等等
% AugGPT如何如何

% 
% DG，DA，black-box KD在我们认知里本质是一个东西，之所以我们这里用KD，我们不仅仅希望大模型做成DAer，还希望把大模型对任务的认知和知识蒸馏出来
% 目前最好的模型是blackbox的大模型，从经济性和效果角度考虑，我们认为最好的KD方法就是黑盒的数据增强方式,which is DG，DA。。。
% 之前的工作只讨论了如何造数据，我们希望

% Data Augmentation (DA) is an effective approach to enhance performance in low-resource scenarios. It expands the number of training samples and increases their diversity.
% The challenges of DA mainly are to maintain the similar concepts and to introduce diversity. For example, using WordNet \cite{miller-1994-wordnet} to replace a word with a synonym, produces limited diversity.
% Pre-trained on a large corpus, LLMs are enhanced by Instruction Learning and Reinforcement Learning from Human Feedback (RLHF) techniques, enabling them to accurately generate diverse texts with specified attributes or labels. These generated texts can then be used to train a base language model under few-shot settings.

Although large language models (LLMs) achieve considerable performance with limited task-specific annotated data \cite{zhang2023controllable, dathathri2020plug, brown2020language,llama1,llama2,yang2023baichuan}, they suffer from the disadvantages of high cost and low speed during inference. Besides, the models of some professional systems are required to perform on a high level for the practical applications, such as coding, math, poem writing, rather than solving diverse tasks. Thus, it is very important to study on efficiently teaching a cheap and small model to learn the professionality of the LLMs \cite{ho2023large, wu2023precedentenhanced}. We consider Knowledge Distillation (KD) as a feasible technique.

\citet{hinton2015distilling} firstly distilled the specific knowledge in an ensemble of models into a single model.
%, combining both of the rich knowledge of heavy model and the advantage of lower runtime cost from small model
Traditional KD methods require to train a teacher model with high-quality annotated data. Facilitated with LLMs, the cost of KD naturally decreases when adopting LLMs as teacher models.

\begin{figure}[t]
    \centering
    \includegraphics[width=\linewidth]{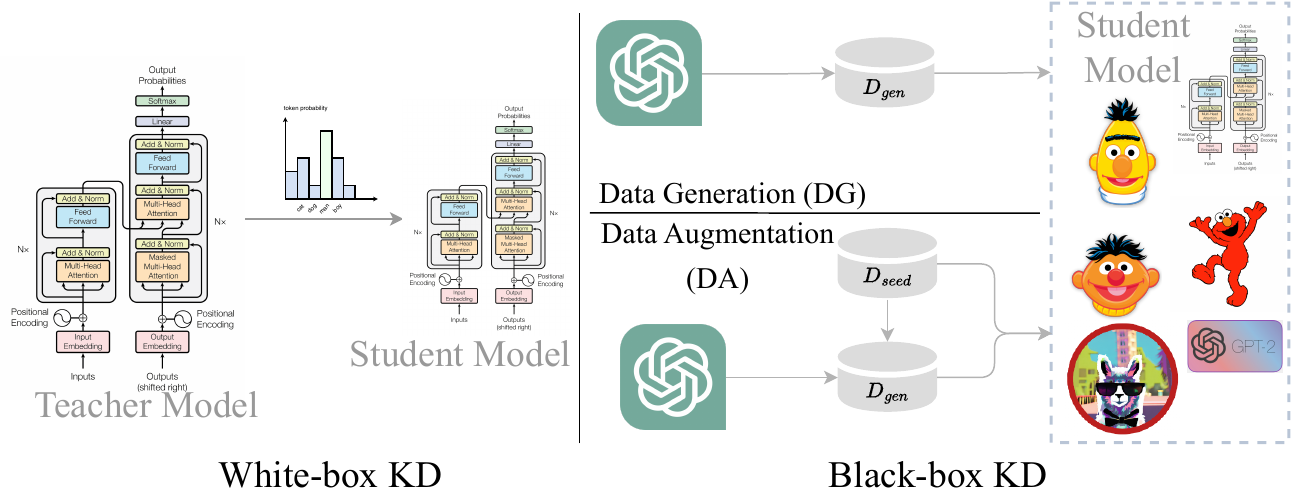}
    \caption{Relationship between KD, DA and DG.}
    \label{fig:kddadg}
\end{figure}

\citet{gu2023knowledge} summarized the two commonly applied categories of KD: \textit{black-box KD}, where only the teacher predictions are accessible, and \textit{white-box KD} \cite{Gou_2021}, where the teacher parameters are available to use. Recently, \textit{black-box KD} has shown promising results in fine-tuning small models on the prompt-response pairs generated by LLM APIs \cite{alpaca, peng2023instruction, lamini-lm}. Comparing with \textit{white-box KD}, \textit{black-box KD}: (1) is less restrictive in terms of structural requirements for teacher models and student model, (2) uses stronger teacher model (such as ChatGPT API), (3) doesn't require the private deployment of teacher model. In \textit{black-box KD}, data serves as the carrier of knowledge since the parameters of teacher model is not available. \citet{ye-etal-2022-zerogen} firstly introduced ZeroGen from the perspective of data-free model-agnostic knowledge distillation in the manner of data generation (DG). Based on DG, \citet{sungen} proposed the noise-robust re-weighting framework. With the superiority of ChatGPT\footnote{https://chat.openai.com/}, \citet{dai2023auggpt} adopted ChatGPT to do Data Augmentation(DA) for text classification, called AugGPT.
Figure \ref{fig:kddadg} illustrates the relationship between DG, DA and KD. The difference between DA and DG is the initial amount of supervised data, which is negligible under 1-shot setting. In this paper, we adopt the concept of "Knowledge Distillation" (KD), because we utilize the LLM not only as a data generator but also to distill knowledge about learning the task, understanding the input text, labeling and evaluating the predictions made by the student model. Currently, the best LLMs are black-box models(such as GPT-4). Considering the cost and efficiency, we adopt \textit{black-box KD} as our distillation approach.

There are two notable limitations in prior works when conducting the process of \textit{black-box KD}: 
\textbf{Under-utilization}, previous studies have regarded LLMs as mere text generators and sentence re-writers, solely relying on their capabilities for text generation and labeling. However, they have neglected the knowledge embedded in the downstream task and the powerful comprehensive ability of LLMs, which may resulting in a hindrance to the quality of the generated text. \textbf{Inflexibility}, prior KD studies have primarily been conducted in an offline and static manner. They construct the entire training data in one go, without considering any dynamic changes that may arise in the status and weaknesses of the student model. Consequently, the generated data often lacks specificity and diversity, limiting its effectiveness in improving the performance of the student model.
% \textbf{Limited scope}, the existing methods are only applied to specific domains or particular tasks.
%Although previous works have achieved success in the field of DA with the help of LLMs, there are great limitations in using DA to improve the few-shot performance of base models on non-specific NLP tasks:
% \begin{itemize}
%     \item The LLMs are viewed as mere text generators and sentence re-writers. Previous studies solely utilize the LLMs capabilities of text generation and labeling, while neglecting the powerful reasoning ability, hindering the quality of the augmented text.
%     \item Previous DA studies are offline and static. They build the whole training data at once, while disregarding changes of the base model's status and weaknesses.
%     Therefore, the augmented data lacks specificity and is limited in effectively improving the performance of the base model.
%     % Previous studies involving Active Learning mainly focus on an offline model or system and apply static methods, while the parameters of our model is continuously changing during training.
%     \item Last but not least, these methods only apply to specific domains or specific tasks.
% \end{itemize}

% Inspired by Dynamic Active Learning, 
In this paper, we propose \textbf{\textsc{EvoKD}}: \textbf{Evo}lving \textbf{K}nowledge \textbf{D}istillation with Large Language Models and Active Learning, in order to address the above limitations.
% Firstly, we introduce the concept of Evolving Knowledge Distillation with Active Learning.
The objective of Active Learning \cite{cohn1996active, literature2009, gentile2022fast} is to optimize the effectiveness of model training by prioritizing the annotation of the most valuable samples.
In line with this, Evolving Knowledge Distillation with Active Learning aims to distil the most informative knowledge that effectively compensate for the weakness of the student model. Moreover, a dynamic teaching strategy is adopted, where the generation of the samples is based on the status of the student model. This dynamic strategy stands in contrast to static strategies that disregard changes in variables.
% why is it evolving
The name draws inspiration from biology, where evolution refers to the change in the characteristics of a species over generations. Changing environmental conditions lead to evolutionary shifts in populations. Similarly, the idea behind Evolving Knowledge Distillation emphasizes the changes of teaching strategy according to the feedback from student, treating it as a dynamic environment over several iterations.

Different from \citet{Moon2019LearnTA, diao2023active}, \textsc{EvoKD} employs LLMs to adapt strategies for generating valuable samples incorporating the concept of Evolving Knowledge Distillation. This approach can mitigating the potential impact of human annotators’ diverse preferences, and providing a stable, cost-effective, and flexible framework.
% Different from previous works using LLMs for DA that augmented all training samples neglecting the base model's status, 
%Active Learning\cite{cohn1996active, literature2009, gentile2022fast} is a technique to find the hard samples with most information, which can be used to compensate for the weakness of the base model. \sihong{to select the most valuable samples for prioritized annotation to maximize the effectiveness of training.} Inspired by Dynamic Active Learning\cite{Moon2019LearnTA}, we continuously traces the status of the base model and provide the most informative examples to a \textbf{dynamic} model, rather than behaves with a static strategy neglecting the change of the variables. \naishan{Different from \citet{Moon2019LearnTA, diao2023active}, we employ LLMs to adjust the strategy to generate the helpful examples in Dynamic Active Learning, which is less expensive and more stable than human labor since LLMs are not influenced by diverse personalities and preferences.}
% \naishan{Compared with vanilla Active Learning\cite{cohn1996active, literature2009, gentile2022fast}, Dynamic Active Learning }

Specifically, \textsc{EvoKD} uses the student model's past performance on selected samples as inputs and prompts the LLM to identify weakness of the student model, based on which, the LLM generates a batch of new sentences, consisting of both challenging and easy samples, along with their corresponding labels. %These newly generated samples are then used to create a batch for the base model. 
The student model is evaluated and subsequently trained on the batch data. The evaluation output provides iterative feedback to the LLM. This batch distillation process is repetitive. The LLM is effectively explored during Weakness Analysis, which demands a meticulous investigation of the data distribution.%Based on these patterns, the LLM generates several new challenging and easy sentences along with their respective labels. We construct a batch for the base model given these new samples and use its output to provide iterative feedback to the LLM, repeating the batch augmentation process.
%\naishan{\textbf{The powerful reasoning ability is fully utilized during pattern reasoning}, which requires careful investigation into the data distribution.}

% \textsc{EvoKD} incrementally expands the training data in batches, with each batch being generated by the online LLM based on its analysis of the weaknesses of the current version's base model. The LLM will adjust its analysis to keep up with changes in parameters and performance of the base model, ensuring that it stays up to date.

% Since \textsc{EvoKD} can be easily leveraged by providing task-specific prompts and a minimal amount of training data, it is an off-the-shelf KD framework with generalization ability.
We performed experiments on five text classification tasks and two NER tasks mainly under 1-shot settings to evaluate the effectiveness of \textsc{EvoKD}. The experiment results demonstrate that \textsc{EvoKD} significantly outperformed the baseline methods. Notably, on text classification datasets, \textbf{\textsc{EvoKD} achieved up to 90\% of the full-shot performance with only 1-shot}. We will release our code for further studies.

% We introduce the preliminaries of large language models, data augmentation and active learning in Section \ref{sec:preliminaries}. The framework of \textsc{EvoKD} is shown in Section \ref{sec:actionda}. The details of the implementation and experimental results are listed in Section \ref{sec:exp}.

Our contributions are three folds, which can be summarized as follows:
\begin{enumerate}
    % \item We investigate an interesting problem on how to fully explore the reasoning capabilities of LLMs to facilitate a more effective data augmentation method.
    %We study a interesting problem on how to fully explore the reasoning ability of LLMs to realize a more effective data augmentation method.
    % \kuangkun{We study a interesting problem on how to explore the reasoning ability of LLMs on data argumentation with improving the quality of the augmented data.}
    \item We introduce the concept of Evolving Knowledge Distillation, which uses dynamically teaching strategy to distill the knowledge about learning the task, understanding the input texts, labeling and evaluating the predictions of student model.
    \item A novel approach called \textsc{EvoKD} is proposed in this paper incorporating the evolving KD and Active Learning, which leverages LLM's potential to comprehend the target task and acquire valuable knowledge.
    % \textsc{EvoKD} can actively chat with online LLMs for dynamic data augmentation. As an off-the-shelf framework, \textsc{EvoKD} offers versatility and can be applied to various NLP tasks across different domains.
    %We propose a novel method, called \textsc{EvoKD}, by actively chatting with online Large Language Model for data augmentation. As an off-the-shelf framework, \textsc{EvoKD} can be generalized to any domain and any tasks.
    % \kuangkun{We propose a novel method, called \textsc{EvoKD}, by Actively Chatting with online Large Language Model for Data Augmentation, inspired by the technique of Dynamic Active Learning.}
    % \item We make full use of the reasoning ability to improve the quality of the augmented data. We propose \textsc{EvoKD}: Actively Chatting with online Large Language Model for Data Augmentation. We also implement two strategies to optimize the iteration process of \textsc{EvoKD}.
    \item Experiments on text classification and NER tasks are conducted under few-shot settings comparing \textsc{EvoKD} with other baselines. \textsc{EvoKD} significantly outperformed all baseline approaches. Notably, \textsc{EvoKD} \textbf{achieved up to 90\% of the full-shot text classification performance with only 1-shot}.
    %We conduct text classification and NER experiments under few-shot settings, comparing \textsc{EvoKD} with other baseline augment methods. \textsc{EvoKD} significantly outperforms other baseline approaches including bare ChatGPT. Notably, we \textbf{achieve up to 90\% of full-shot text classification performance with only 1-shot}.
\end{enumerate}

\begin{figure*}
    \centering
    \includegraphics[width=\linewidth]{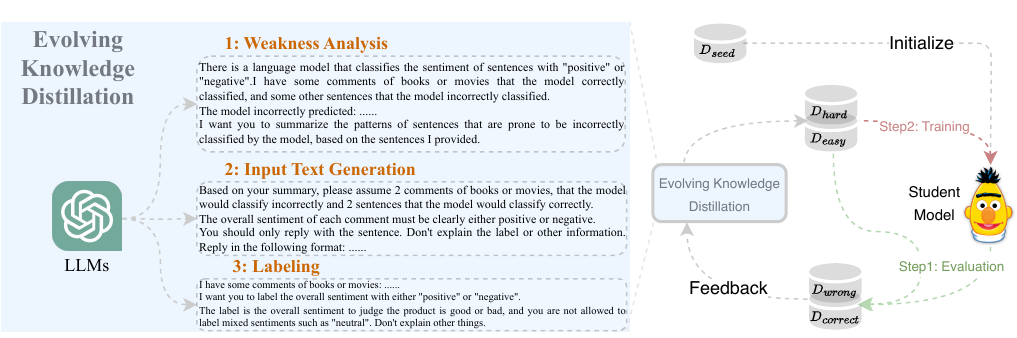}
    \caption{Framework of \textsc{EvoKD}. The initial student model is trained using the few-shot training data. Then, both correct and wrong samples are identified via ``Evaluation'' step. Iteratively, the identification results are used to distill the new samples. For Evolving Knowledge Distillation, the process begins by prompting the LLM to analyse the weakness of the student model, given the correct and wrong samples. Based on the weakness, the LLM is required to generate a set of challenging and easy samples, which are collected to construct a batch data. The batch data firstly evaluates the student model to obtain the next feedback, then the student model is trained on the batch data.
    % For Evolving Active Learning, the process begins by prompting the LLM to summarize the patterns in the weaknesses of the base model, given the correct and wrong samples. Based on the pattern, the LLM is required to generate a set of challenging and easy samples, which are collected to construct a batch data. The base model inferences on the batch data to identify the correct and wrong samples for the next iteration, and updates its parameters by training on the batch data.
    }
    %\caption{Framework of \textsc{EvoKD}. We use the few-shot training data to train the initial base model. Then the correct samples and wrong samples are identified. For Evolving Active Learning, we start from prompting LLM to summarize the pattern of the weakness of the base model given the correct and wrong samples. Based on the pattern, the LLM is asked to generate several hard and easy samples, which are collected to construct a batch data. The base model inferences on the batch data to identify the correct and wrong samples for the next iteration, and updates its parameters by training on the batch data.}
    \label{fig:framework}
\end{figure*}

\section{Related Work}\label{sec:preliminaries}

\subsection{Large Language Model}

Large Language Models (LLMs) have recently demonstrated remarkable superiority in multiple traditional NLP tasks.
%Large Language Models (LLMs) have shown significant superiority in the field of several traditional NLP tasks recently.
% The success of LLMs benefits from the following aspects: 1) LLMs are firstly pre-trained on a large number of collected corpus in the manner of auto-regression, which predicts the probability of the next token given the previous tokens. The common sense knowledge and factual knowledge in the pre-training corpus can be memorized in the pairs of keys and values in the parameters \cite{geva-etal-2021-transformer}. Thus LLMs perform amazing  ability to understand and generate texts. 2) Instruction Learning \cite{brown2020language, gpt2, fan2018learning}. In order to prompt LLMs to generate proper answers to questions, rather that just the continuation of the inputs, researchers train LLMs on several task-specific instructions, with the corresponding labels as outputs. In this way, users are replied with direct answers by instructions, without extra trials of different prompts. 3) Reinforcement Learning from Human Feedback (RLHF) \cite{ziegler2020finetuning,lambert2022illustrating}, which is a technique that trains a reward model directly from human feedback and uses the model as a reward function to optimize an agent's behaviour or response using reinforcement learning through an optimization algorithm like Proximal Policy Optimization \cite{schulman2017proximal}.
\citet{brown2020language} proposed GPT-3, increasing the size of model parameters to 175B. \citet{chowdhery2022palm} trained a 540B parameter, densely activated Transformer language model, named as PaLM. PaLM achieved SOTA few-shot learning results on hundreds of NLU and NLG taks. \citet{chung2022scaling} released Flan-T5, they found that instruction fine-tuning with the scaling of the number of tasks, the model size and fine-tuning on chain-of-thought data dramatically improves performance.
% \citet{touvron2023llama} pre-trained LLaMa using publicly available datasets exclusively, without resorting to proprietary and inaccessible datasets. \citet{alpaca} fine-tuned Alpaca from a 7B LLaMA model on 52K instruction-following data generated by Self-Instruction \cite{wang2022selfinstruct}.

% \subsection{Data Augmentation and Data Generation under Low-Resource Scenario}
\subsection{Black-Box KD}

Few-shot learning is considered as a more practical setting than full-shot, although some methods are developed to improve the performance of models \cite{Wang_Cui_Wang_Pei_Zhu_Yang_2017}.
Data augmentation is a widely employed approach to improve the performance by expanding the size of the training dataset and increasing the diversity of samples.
Common data augmentation approaches for NLP tasks include deleting, inserting random characters, and substituting words with synonyms. However, these methods have obvious limitations such as reduced text fluency and limited word diversity.

% With the rise of large language models, the performance of data augmentation has significantly improve, since LLMs can generate fluent and diverse sentences.
Model-based methods are more efficient currently.
\citet{edwards2023guiding} used GPT-2 to generate artificial training instances with domain expert selection in order to improve classification performance.
\citet{wei-etal-2021-shot} explored a technique particularly suitable for few-shot, highly-multiclass text classification setting. To further boost performance, they also presented a simple training strategy called curriculum data augmentation, which leverages curriculum learning by first training on only original examples and then introducing augmented data as training progresses. \citet{dai2023auggpt} proposed a text data augmentation approach based on ChatGPT, named AugGPT. AugGPT rephrases each sentence in the training samples into multiple conceptually similar but semantically different samples.

Data Generation is slightly different with DA, as it has no initial seed data. 
ZeroGen \cite{ye-etal-2022-zerogen} is a flexible and efficient zero-short learning method, also provides insights from the perspective of data-free model-agnostic knowledge distillation. \citet{sungen} proposed a novel noise-robust re-weighting framework SunGen to automatically construct high-quality data for zero-shot
classification problems. \citet{ubani2023zeroshotdataaug} investigated the use of data obtained from prompting a large generative language model, to generate synthetic training data for few-shot learning. \citet{tang2023does} proposed to generate a vast quantity of high-quality synthetic data with labels utilizing ChatGPT and fine-tuning a local model for the downstream task. They prompted ChatGPT to extract structured information from unstructured healthcare texts, with a focus on biological named entity recognition and relation extraction.

% Different from previous works, we prompt a LLM for DA to solve non-specific NLP tasks in open domain. We thoroughly exploit the reasoning abilities of LLMs actively, rather than just regarding them as sentence re-writers. The LLM plays a dual role in both analyzing outputs and generating new samples.

\subsection{Active Learning}

Active Learning involves reducing the amount of labeled data needed to learn a target concept by strategically querying the annotator for labels of the most informative examples \cite{yuan2020cold, angluin1988queries, sener2017active, settles2009active}.
% This approach enables the concept to be learned with a reduced number of examples.
%Active learning is the task of reducing the amount of labeled data required to learn the target concept by querying the user for labels for the most informative examples so that the concept is learnt with fewer examples.
\citet{diao2023active} proposed Active-Prompt, to adapt LLMs to different tasks with task-specific example prompts which are annotated with human-designed CoT reasoning, and they determined which questions are the most important and helpful ones to annotate from a pool of task-specific queries. \citet{wang-etal-2021-want-reduce} proposed an active labeling strategy
to have humans re-annotate data labeled by GPT-3 \cite{brown2020language}
with the lowest confidence scores, to reduce the noise in the labeled data from GPT-3. There are also other approaches to select the instances to be labeled \cite{schumann2019active, ren2021survey}.

Standard Active Learning operates by utilizing a pool of unlabeled data, from which annotators select the most informative samples for annotation, thereby reducing labeling costs. However, our experimental setup differs significantly. Our primary focus revolves around the task of knowledge distillation, wherein we compare our approach against other baselines about KD. Unlike traditional Active Learning scenarios, our framework does not involve human annotators or rely on an unlabeled data pool. Instead, we only draw inspiration from the core motivation of Active Learning, which involves identifying the most informative instances. In our framework, these instances are generated by LLM, further distinguishing it from standard Active Learning tasks.

\section{\textsc{EvoKD}} \label{sec:actionda}

In this Section, we introduce the framework of \textsc{EvoKD}, as shown in Figure \ref{fig:framework}.

In few-shot learning, only a limited amount of training data is available initially. We begin with 
$m$ training samples and use them to train a student model, denoted as $model^0$. Some samples are predicted incorrectly by $model^0$ and are denoted as $D_{wrong}^0$, while others are predicted correctly and are denoted as $D_{correct}^0$.

We use the LLM to perform knowledge distillation on the incorrectly predicted or classified samples to investigate the student model's weaknesses on specific tasks, which is an active approach. As the parameters of the student model changes, its performance and weakness both change accordingly.
%the model's performance changes, and so does the pattern as well as the augmented batch samples for training.

\textsc{EvoKD} performs knowledge distillation iteratively. For the $i$-th iteration, $D_{wrong}^{i-1}$, $D_{correct}^{i-1}$ and $model^{i-1}$ are given, where $D_{wrong}^{i-1}$ is the subset of samples where $model^{i-1}$ has the worst performance, and $D_{correct}^{i-1}$ is the subset of samples with the best performance. The online LLM is fed the student model's performance and it is asked to generate several samples in a conversational manner. %We feed the base model's performance to the online LLM and ask it to generate several samples in a conversational manner.
Formally, the LLM is asked to propose $ \lfloor \frac{b}{2} \rfloor$ easy samples, denoted as $D_{easy}^i$, and $\lceil \frac{b}{2} \rceil$ hard samples, denoted as $D_{hard}^i$, given $D_{wrong}^{i-1}$ and $D_{correct}^{i-1}$, where $b$ is the batch size. Then $D_{easy}^i$ and $D_{hard}^i$ are concatenated to construct the $i$-th batch $D^i$, with which, $model^{i-1}$ is updated to $model^i$. Additionally, the current student model is evaluated with $D^i$. The teacher LLM is then instructed to generate the new batch data $D^{i+1}$ given the status of current model performance.

% \jiang{There are two reasons for the inclusions of $D_{correct}$ in the chatting inputs and $D_{easy}$ in the chatting outputs:}
There are reasons for the inclusions of $D_{correct}$ in the chatting inputs and $D_{easy}$ in the chatting outputs respectively:
%Including both $D_{easy}$ and $D_{correct}$ in the chatting inputs and outputs respectively improves the performance, which is proved in the ablation study. We list the reasons as follows:
\begin{itemize}
    \item \textbf{Why include $D_{correct}$ to construct inputs?} The correctly predicted samples help the LLM analyse the student model's weaknesses. Including the correctly predicted samples makes the weakness more apparent than using only the incorrectly predicted samples.
    \item \textbf{Why use $D_{easy}$ to train?} If we only use the challenging samples as training data, the sample distribution learned will be biased, leading to the problem of catastrophic forgetting. The student model may perform increasingly poorly on originally easy samples. Therefore, we enable LLM to generate both $D_{easy}$ and $D_{hard}$ based on weakness, thereby preventing the base model from forgetting previous knowledge and falling into local optimum caused by biased distributions.
    % There is a risk that the LLM may guess wrong patterns, and the hard cases it outputs may actually be easy or not very hard samples for the base model. In this case, the LLM has little chance of finding the pattern in the next iteration and changing its conclusion, due to the lack of the base model's performance on really hard samples.
\end{itemize}
In fact, the above operations indeed improve performance, as demonstrated by the ablation study (in SubSection \ref{sec:exp pipeline}).

The batch $D^i$ is used in two ways:
1) The previous checkpoint $model^{i-1}$ is evaluated on $D^i$ , and the real incorrect samples $D_{wrong}^i$ and correct samples $D_{correct}^i$ are identified based on the metric. The performance is then described in the prompt, which is fed to the LLM in the next iteration.
2) $D^i$ also serves as the knowledge carrier to update the student model from $model^{i-1}$ to $model^i$. The model checkpoint $model^n$ after $n$ iterations is the final objective.

\subsection{Evolving Knowledge Distillation with LLM}\label{subsec:chat pipeline}

Evolving Knowledge Distillation with LLM aims to dynamically provide the most informative knowledge for the student model. In \textsc{EvoKD}, the LLM is utilized to adapt teaching strategies to generate the beneficial samples. To enhance the results of Evolving Knowledge Distillation, we have subdivided the batch generation process into three sub-steps:
%Evolving Active Learning aims to dynamically find the most informative samples for the evolving base model.
%In \textsc{EvoKD}, the online LLM is employed to adapt DA strategies to generate the valuable samples. To enhance the results of Evolving Active Learning, we have subdivided the batch generation process into three sub-steps:
\begin{enumerate}
    \item \textbf{Weakness Analysis.} The LLM analyses the weakness of the student model by identifying the pattern of the sentences that are likely to be incorrectly predicted by the student model. The pattern string is used both in generating new samples and explaining the generated samples. If required, \textsc{EvoKD} also enables human intervention to adjust the pattern string, thus influencing subsequent generations.%\textsc{EvoKD} even allows extra interference from human to adjust the pattern, thereby influencing the subsequent generation.
    \item \textbf{Input Text Generation.} The LLM is prompted to create input texts according to the identified weakness. LLMs exhibit advanced reasoning capabilities, enabling them to generate text based on attributes inferred from weakness. %following the pattern. Leveraging the advanced reasoning capabilities of LLMs, they can generate texts based on the attributes inferred from the pattern.
    \item \textbf{Labeling.} The LLM is prompted to label each generated sentence in a new conversation. We separate the text generation and labeling processes to mitigate the risk of intentional mislabeling by the LLM. As in the conversation of Input Text Generation, we requested the LLM to create challenging samples, if the LLM simultaneously generates text and labels, it may deliberately mislabel the samples to induce the student model to make mistakes. We conducted ablation study to prove the effectiveness of separating text generation and labeling, which is shown in SubSection \ref{sec:exp pipeline}.%We ask the LLM to label the category for each generated sentence in a new conversation. the LLM may potentially mislabel the categories \naishan{on purpose} in order to induce the base model to make mistakes, particularly when we prompt the LLM to simultaneously generate text and labels, which is not what we expect. We conducted ablation study to improve the effectiveness of separating text generation and labeling, which is shown in SubSection \ref{sec:exp pipeline}.
\end{enumerate}

% We show an example of the interaction with the online LLM to solve the task of sentiment classification using \textsc{EvoKD} in Appendix \ref{sec:chat example}.

% We find that dividing the chat process into the aforementioned sub-steps improves the quality of the augmented samples.
% The manner of ``generate step by step'' also alleviates ChatGPT's the pressure on reasoning and understanding.

\begin{algorithm}[tb]
	\caption{\textsc{EvoKD}} 
	\label{alg1} 
	\begin{algorithmic}[1]
		\REQUIRE $D^0, num\_steps, chat, review$
            \STATE Initialize $model^0$
            \STATE $D_{wrong}^0, D_{correct}^0 \gets \mathbf{Identify}(model^0, D^0)$
            \STATE $model^0 \gets$ train $model^0$ on $D^0$
		\STATE $i \gets 0$
            \STATE $step \gets 0$
            \STATE $history \gets \{ D^0\}$
		\WHILE{$step < num\_steps$} 
                \STATE $step \gets step + 1$
    		\IF{$step$ $\%$ $review = 0$} 
                \STATE train $model^i$ on $history$
                \STATE update $step$
    		\ELSIF{$step$ \% $chat = 0$}
                \STATE $i \gets i + 1$
                \STATE {\small $D_{hard}^{i}, D_{easy}^{i}\gets\ \mathbf{LLM}(D_{wrong}^{i-1}, D_{correct}^{i-1})$}
                \STATE $D^i \gets \{ D_{hard}^{i}, D_{easy}^{i} \}$
                \STATE Add $D^{i}$ to $history$
                \STATE {\small $D_{wrong}^{i}, D_{correct}^{i} \gets \mathbf{Identify}(model^{i-1}, D^i)$}
                \STATE $model^i \gets$ train $model^{i-1}$ on $D^i$
                \ELSE
                \STATE train $model^i$ on $D^i$
		      \ENDIF 
		\ENDWHILE
        \RETURN $model^i$
	\end{algorithmic} 
\end{algorithm}

\subsection{Strategies to Improve Effectiveness}

% The training and inference of the base model are executed locally. The primary hindrance to reducing training time is the interaction with ChatGPT. 
% Therefore we suggest two strategies to reuse the samples generated by ChatGPT.

We suggest two strategies to reuse the samples generated by the LLM. 
% \jiang{\sout{As demonstrated in the experiments(where???), these two strategies do improve effectiveness.}} %which improve effectiveness, demonstrated by the experiments.
\paragraph{Repeat Batch} We train the student model on the same batch for several steps. Because training on a batch of samples for a single step has little impact to the parameters and performance of the student model, and its prediction will likely be similar to the previous iteration.
\paragraph{Review History} We store all the generated samples in a global cache. Then, at regular intervals, batches of samples are fetched from the cache to train the model. We find that training on the historical samples enables the model to recall previous knowledge and avoid making mistakes on the same patterns.

% With the above two strategies, not only the cost of communication decreases, the base model also reviews previous trained batches, which prevents it from making the similar mistakes
The pseudocode of \textsc{EvoKD} is shown in Algorithm \ref{alg1}.
%To make the pipeline clearer, we program \textsc{EvoKD} in Algorithm \ref{alg1}.
% The process of \textsc{EvoKD} saving the cost is shown in Algorithm \ref{alg1}.

\subsection{Initialization}
If we execute \textsc{EvoKD} with a randomly initialized student model, the initial pattern may be arbitrary and unnecessary to analyze, or it could potentially hinder the student model's ability to comprehend texts with diverse attributes. To address this issue, we introduce an optional strategy: AugGPT is incorporated in the first few epochs, before running \textsc{EvoKD} in the subsequent epochs.

\section{Experiments}\label{sec:exp}

\begin{table}[th]
    \centering
    \scriptsize
    \begin{tabular}{m{1.2cm}<{\centering}m{1cm}<{\centering}m{1cm}<{\centering}m{0.7cm}<{\centering}m{0.7cm}<{\centering}m{0.7cm}<{\centering}}
    \toprule
        Task & Dataset & Language & Label Num & Train Num & Test Num\\
        \midrule
        \multirow{5}{*}{Classification} & Amazon & English & 2 & 50000 & 50000 \\
        & IMDB & English & 2 & 25000 & 25000\\
         & Inshorts & English & 5 & 2999 & 407\\
         & Toutiao News & Chinese & 14 & 10000 & 950\\
         & CAIL2019 divorce & Chinese & 3 & 9876 & 1200\\
         \midrule
         \multirow{2}{*}{NER} & CoNLL03 & English & 3 & 14041 & 3453\\
         & CoNLL04 & English & 3 & 922 & 288 \\
         \bottomrule
    \end{tabular}
    \caption{Details of the datasets.}
    \label{tab:data}
\end{table}

\begin{table*}[thb]
    \centering
    \scriptsize
    % \fontsize{8pt}{11pt}\selectfont
    % \rmfamily
    % \begin{tabular}{m{1.9cm}m{2.2cm}<{\centering}m{2.2cm}<{\centering}m{2.2cm}<{\centering}m{2.2cm}<{\centering}m{2.2cm}<{\centering}c}
    \begin{tabular}{m{1.6cm}|m{2cm}<{\centering}m{2.0cm}<{\centering}m{2.0cm}<{\centering}m{2.0cm}<{\centering}m{2.0cm}<{\centering}|c}
    \toprule
        \multirow{2}*{\normalsize Method}& \multicolumn{3}{c}{\normalsize English} & \multicolumn{2}{c|}{\normalsize Chinese} & \multirow{2}*{\normalsize AVG} \\
         & Amazon & IMDB & Inshorts & TouTiao & CAIL2019 & \\
    \midrule
    Full Shot & 0.9480	& 0.9495&	0.9705	&0.8495&0.9683 & 0.9372\\
    \midrule
    No Augment & 0.6030 $\pm$ 0.0880 & 0.5833 $\pm$ 0.0853 &0.6408 $\pm$ 0.1528 & 0.3638 $\pm$ 0.0737&	0.4422 $\pm$ 0.0604 & 0.5266\\
    % Swap Char & 0.6237 $\pm$ 0.0904	& 0.6108 $\pm$ 0.0659 & 0.6762 $\pm$ 0.1628	& 0.3848 $\pm$ 0.0843	& 0.5683 $\pm$ 0.1192 & 0.5728\\
    % Delete Char & 0.6314 $\pm$ 0.1011&	0.5972 $\pm$ 0.0776& 0.6663 $\pm$ 0.1588	&0.3554 $\pm$ 0.0842 &	0.5357 $\pm$ 0.1131 & 0.5559\\
    % Insert Char & 0.5618 $\pm$ 0.1221	&0.6189 $\pm$ 0.0599 & 0.6776 $\pm$ 0.1669 &	0.3808 $\pm$ 0.0729	& 0.6120 $\pm$ 0.1238 & 0.5702\\
    % Replace Word & 0.6314 $\pm$ 0.0838 &0.5608 $\pm$ 0.1401 &	0.6290 $\pm$ 0.1699	& 0.3735 $\pm$ 0.0668 &	0.3933 $\pm$ 0.0319 & 0.5176\\
    % Insert Word & 0.6076 $\pm$ 0.0843 & 0.5628 $\pm$ 0.1429 &	0.6295 $\pm$ 0.1868 & - & - & -\\
    EDA & 0.6314 $\pm$ 0.0838 & 0.6189 $\pm$ 0.0599 & 0.6776 $\pm$ 0.1669 & 0.3848 $\pm$ 0.0843 & 0.6120 $\pm$ 0.0319 & 0.5849\\
    ZeroGen & 0.7054 $\pm$ 0.1134&	0.5087 $\pm$ 0.2204&	0.8334 $\pm$ 0.0429	&0.6442 $\pm$ 0.0369	&0.7620 $\pm$ 0.1118 & 0.6907\\
    SunGen & 0.6257 $\pm$ 0.1288 &	0.5769 $\pm$ 0.0521 &	0.8103 $\pm$ 0.0456 &	0.2533 $\pm$ 0.0827 &	0.8305 $\pm$ 0.1020 & 0.6193 \\
    Gradual & 0.5826 $\pm$ 0.0771&	0.6857 $\pm$ 0.0109 &	0.7608 $\pm$ 0.0144 &- &- & -\\
    AugGPT & 0.6234 $\pm$ 0.1712	&0.6903 $\pm$ 0.0788 & 0.7902$\pm$ 0.0759	&	0.6514 $\pm$ 0.0459 &	0.7122 $\pm$ 0.1117 & 0.6935\\
    \midrule
    \textsc{EvoKD} & \textbf{0.8425 $\pm$ 0.0317} &	\underline{0.7982 $\pm$ 0.0565}	& \underline{0.8516 $\pm$ 0.0257} &	\underline{0.6874 $\pm$ 0.0199} &	\textbf{0.9148 $\pm$ 0.0411} & \underline{0.8189}\\
    \quad +Init & \underline{0.8403 $\pm$ 0.0240} &	\textbf{0.8359 $\pm$ 0.0272} & \textbf{0.8688 $\pm$ 0.0167} &	\textbf{0.7112 $\pm$ 0.0237} & \underline{0.9137 $\pm$ 0.0355} & \textbf{0.8340}\\
    \bottomrule
    \end{tabular}
    \vspace{1em}
    \caption{Experiment results under 1-shot text classification. We use ``No Augment'' to denote the 1-shot performance without knowledge distillation, and ``+Init'' means that in the initial epochs, using AugGPT to initialize the student model. The best results are highlighted in bold, and the second best results are underlined. ``AVG'' denotes the average performance over all datasets.}
    %in the first few epochs direct text augmentation with ChatGPT is applied to initialize the base model, and then in the remaining epochs the base model is trained with \textsc{EvoKD}. The best results are highlighted in bold, and the second best results are underlined. ``AVG'' denotes the average performance over all datasets.}
    \label{tab:1-shot cls exp}
\end{table*}

In this Section, we conduct experiments on 5 text classification datasets under 1-shot setting (SubSection \ref{sec:exp main}), and investigate the performance when the size of training data increases (SubSection \ref{sec:exp fewshot}). We also explore the effectiveness of \textsc{EvoKD} on NER tasks (SubSection \ref{sec:exp gen}). Finally, we conduct the ablation study (SubSection \ref{sec:exp pipeline}).

\subsection{Implementation Details}

We utilize gpt-3.5-turbo-0301 to implement the conversation with the LLMs, and report the mean F1 value and standard deviation with random seeds ranging from 1 to 5 for each setting.
We adopt BERT-base as the student model for English datasets and Chinese-BERT-base for Chinese datasets. We set the learning rate to 2e-5 and batch size to 8. We set the clip gradient norm to 2. For few-shot experiments, we train the student model with 10 epochs, and the total number of steps in each epoch is set to 1250. As we have no available validation set in the practical few-shot scenario, we directly test the final checkpoint of the student model. For few-shot sampling from full training data, we follow \cite{dai2023auggpt}, \citet{uie} and \citet{usm}.
To train \textsc{EvoKD}, we set $chat$ in Algorithm \ref{alg1} to 40, and $review$ to 50. And the threshold to identify the correct and wrong cases is set to 0.95.

\subsection{Datasets}

Our experiments encompass two tasks: Text Classification and Named Entity Recognition (NER). 
For Text Classification, we use Amazon \citelanguageresource{amazon}, IMDB \citelanguageresource{imdb} and Inshorts-News\footnote{https://github.com/kishanpython/Inshorts-News-Data-Collection}, TouTiao-News\footnote{https://github.com/BenDerPan/toutiao-text-classfication-dataset} and CAIL2019\footnote{https://github.com/china-ai-law-challenge/CAIL2019/}. Amazon and IMDB are datasets for product reviews, which express either a positive or negative sentiment. Inshorts and TouTiao are about news categories classification. We remove the ambiguous news categories. The track we choose from CAIL2019 involves identifying attributes from sentences relative to divorce events. We only keep the categories ``have children after marriage'', ``joint debts'' and ``joint property'' in order to clearly distinguish between the attributes.
For NER, we use CoNLL03 \citelanguageresource{conll03} and CoNLL04 \citelanguageresource{conll04}. We remove the entity types ``other'' and ``miscellaneous'' as they have no specific meanings.

We list the details of the datasets in Table \ref{tab:data}.

\subsection{Baselines}

As we discussed in Section \ref{sec:intro}, both of DG and DA can be applied in \textit{black-box KD}. Several previous studies of DG and DA are included as baselines:

%We have not included \citet{dai2023auggpt} as one of the baselines because the code is not publicly available \naishan{at the time of the writing}. Instead, we implement a new DA method using ChatGPT, which generates sentences in a variety of styles and contents. We list the baselines as follows:
\paragraph{EDA}
We adopt several rule-based approaches for Easy Data Augmentation(EDA), including 1) Swap Char \cite{belinkov2018synthetic}, 2) Delete Char, 3) Insert Char, 4) Replace Word \cite{ma2019nlpaug}, 5) Insert Word. We employ nlpaug \cite{ma2019nlpaug} for English and nlpcda\footnote{https://github.com/425776024/nlpcdaf} for Chinese to implement the rule-based methods. The baselines "Swap Word" and "Replace Word" both use synonym words, which are generated by stronger, model-based methods. For details, we adopt bert-base-uncased from HuggingFace as the model for synonyms. We set $change\_rate$ as 0.3 in nlpcda, and the others are default. The highest F1 score among above methods is reported.
% \paragraph{Swap Char}
% Swapping two random characters in the original sentences \cite{belinkov2018synthetic}.
% \paragraph{Delete Char}
% Randomly deleting a character from the original sentence.
% \paragraph{Insert Char}
% Inserting a random character into a random position of the origin sentence.
% \paragraph{Replace Word}
% Replacing a random word in the original sentence with a synonym word which is consistent with the context\cite{ma2019nlpaug}. For a Chinese sentence, we replace a whole phrase.
% \paragraph{Insert Word}
% Inserting a word into the original sentence, and the inserted word is consistent with the context of the sentence. For Chinese datasets, we didn't implement this approach.
\paragraph{AugGPT}

\citet{dai2023auggpt} introduced AugGPT for few-shot learning to solve text classification. As the code and prompt is publicly unavailable at the time of writing, we write customized prompt for each dataset.

% Given an example, we prompt ChatGPT to generate a new sentence. Firstly, we introduce the task to ChatGPT, outlining the inputs, outputs, and label definitions. Following this, we ask ChatGPT to generate a sample according to the given example. The new sentence generated by ChatGPT does not need to have the same topic or content as the original sentence, but it must include the same specific attribute or label. In this way, ChatGPT can generate sentences in a variety of styles, rather than simply serving as a sentence rewriter. 
% % We list some showcases in Appendix \ref{sec:ChatGPT Augmentation Example}.

% We have excluded \citet{dai2023auggpt} as one of the baselines due to the unavailability of its code at the time of writing.

\paragraph{ZeroGen}

ZeroGen \cite{ye-etal-2022-zerogen} provides insights from the perspective of data-free model-agnostic knowledge distillation. For fairness, we set the same augmented data size for all methods under 1-shot, and adopt gpt-3.5-turbo-0301 to do augmentation with seed samples as context.

\paragraph{SunGen}

\citet{sungen} proposed a novel noise-robust re-weighting framework SunGen. SunGen shares the same data setting with ZeroGen.

\paragraph{Gradual Curriculum}

\citet{wei-etal-2021-shot} explored a technique particularly suitable for few-shot, highly-multiclass text classification setting. As the official implementation depends on EDA package which doesn't supports Chinese, so we only report the results on English datasets. This baseline is denoted as \textbf{Gradual} in the following experiments.

\subsection{1-Shot Classification}\label{sec:exp main}

% We conduct experiments on the classification datasets with 5 different seeds and report the average f1 and standard deviation. We compare the results of \textsc{EvoKD} with 1-shot baseline text augmentations, bare ChatGPT augmentation and full-shot performance in Table \ref{tab:1-shot cls exp}.
We can observe from Table \ref{tab:1-shot cls exp} that,
model-based methods exhibits notable superiority over EDA.
% Baselines that focus on fine-grained text content insertion, deletion, and replacement only yield marginal improvements.
% Through our prompt text, ChatGPT may introduce thematic and grammatical diversity into samples, thereby enhancing the performance of the base model.
SunGen has a clear disadvantage in the news classification of TouTiao. We observed its sample weights and found that SunGen gives a relatively large weight to the $story$ and $travel$ categories, while the weights for other categories are small. Through analysis, we discovered that due to the low noise in the samples generated by GPT-3.5, SunGen's advantage cannot be fully utilized. Under this setting, when there are multiple label categories, SunGen's performance can be unstable and unsatisfactory.

Our proposed method, \textsc{EvoKD}, exhibits \textbf{better performance} than all baselines on all classification datasets. 
Compared with AugGPT, \textsc{EvoKD} achieved the highest absolute improvement in Amazon sentiment classification at 21.91\% and the lowest in TouTiao news classification at 3.60\%. It is worth mentioning that \textsc{EvoKD} demonstrates a \textbf{higher stability} compared with AugGPT. For instance, in the case of Amazon, \textsc{EvoKD} reduces the deviation from 0.1712 to 0.0317, resulting in a relative reduction of 81.48\%. The stability of \textsc{EvoKD} could be attributed to its active analysis of the student model, which effectively mitigating the impact caused by the distribution randomness of the seed samples.%We also notice that, \textsc{EvoKD} exhibits a lower deviation than bare ChatGPT. To illustrate, in the case of Amazon, \textsc{EvoKD} reduces the deviation from 0.1712 to 0.0317, representing a relative decrease of 81.48\%. The stability of \textsc{EvoKD} comes from its active analysis of the base model, which addresses the influence caused by the distribution bias of seed samples.

``\textsc{EvoKD} + Init'' involves distilling the training data in the first few epochs using AugGPT for initialization, then training the student model with \textsc{EvoKD} in the subsequent epochs. The results are shown in the last row of Table \ref{tab:1-shot cls exp}. 
``\textsc{EvoKD} + Init'' \textbf{achieved as high as 90\% of the full-shot performance with only 1-shot} on most of the text classification datasets, such as CAIL2019 and Inshorts news classification. In the case of TouTiao news classification, which involves 14 categories, \textsc{EvoKD} with initialization achieved 84\% of the full-shot F1 under the 1-shot setting. These results highlight \textsc{EvoKD}'s ability to effectively employ LLMs for actively analyzing and generating difficult samples in 1-shot scenarios. By harnessing the capabilities and knowledge of LLMs, \textsc{EvoKD} showcases its effectiveness in knowledge distillation.%The results demonstrate the efficacy of leveraging ChatGPT to actively analyze and augment difficult samples in few-shot scenarios.

We also notice that on the legal dataset, CAIL2019, \textsc{EvoKD} outperforms the baselines by at least 8 percents, which indicates the effectiveness of \textsc{EvoKD} under professional domains.

\begin{table*}[thb]
    \centering
    \scriptsize
    \begin{tabular}{lm{0.7cm}<{\centering}m{2.0cm}<{\centering}m{2.0cm}<{\centering}m{2.0cm}<{\centering}m{2.0cm}<{\centering}m{2.0cm}<{\centering}c}
    \toprule
        \multirow{2}*{\normalsize Method} & \multirow{2}*{\normalsize Shot} & \multicolumn{3}{c}{\normalsize English} & \multicolumn{2}{c}{\normalsize Chinese} & \multirow{2}*{\normalsize AVG}\\
         & & Amazon & IMDB & Inshorts & TouTiao & CAIL2019 \\
    \midrule
    Full Shot & Full & 0.9480	& 0.9495&	0.9705	& 0.8495 & 0.9683 & 0.9372\\
    \midrule
    No Augment & \multirow{4}{*}{3} & 0.6704 $\pm$ 0.0446&	0.6269 $\pm$ 0.0720&	0.8182 $\pm$ 0.0429	& 0.6623 $\pm$ 0.0254&	0.7792 $\pm$ 0.0403 & 0.7114\\
    AugGPT & & 0.7574 $\pm$ 0.0808	&0.7266 $\pm$ 0.1479	&\textbf{0.8639 $\pm$ 0.0189}	& 0.6985 $\pm$ 0.0300 & 0.9000 $\pm$ 0.0606 & 0.7893\\
    \textsc{EvoKD} & & \textbf{0.8392 $\pm$ 0.0066} &	\textbf{0.7873 $\pm$ 0.0574}	&	0.8501 $\pm$ 0.0335	&	\textbf{0.7328 $\pm$ 0.0120}	&\underline{0.9263 $\pm$ 0.0150} & \textbf{0.8271}\\
    \quad +Init & & \underline{0.8270 $\pm$ 0.0472} &	\underline{0.7808 $\pm$ 0.0681}	&	\underline{0.8561 $\pm$ 0.0359}	&	\underline{0.7259 $\pm$ 0.0122}	& \textbf{0.9383 $\pm$ 0.0124} & \underline{0.8256}\\
    \midrule
    No Augment & \multirow{4}{*}{5} & 0.7159 $\pm$ 0.0421 &	0.7536 $\pm$ 0.0388 &	0.8693 $\pm$ 0.0292&	0.7352 $\pm$ 0.0096&	0.8937 $\pm$ 0.0245 & 0.7935\\
    AugGPT & & \underline{0.8333 $\pm$ 0.0215}	 & \underline{0.7858 $\pm$ 0.0728} & 0.8717 $\pm$ 0.0249 & 0.7248 $\pm$ 0.0256 & \underline{0.9413 $\pm$ 0.0067} & 0.8314\\
    \textsc{EvoKD} & & 0.8228 $\pm$ 0.0301 &	\textbf{0.8485 $\pm$ 0.0463}	&	\textbf{0.8826 $\pm$ 0.0179}	&	\textbf{0.7474 $\pm$ 0.0196} &	0.9307 $\pm$ 0.0138 & \textbf{0.8464}\\
    \quad +Init & & \textbf{0.8682 $\pm$ 0.0130} &	0.7729 $\pm$ 0.1607	& \underline{0.8767 $\pm$ 0.0081} & \underline{0.7445 $\pm$ 0.0118} &	\textbf{0.9460 $\pm$ 0.0099} & \underline{0.8417}\\
    \bottomrule
    \end{tabular}
    \vspace{1em}
    \caption{Experiment results under Few-Shot text classification.}
    \label{tab:few-shot cls exp}
\end{table*}

\subsection{Efficiency of EvoKD}

\begin{figure}[thb]
\centering  
\includegraphics[width=\linewidth]{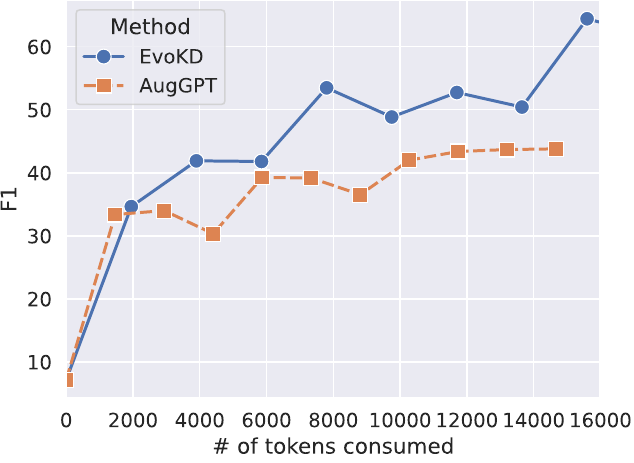}
\caption{F1 versus the number of tokens used during training.}
\label{fig:numofcall}
\end{figure}

Figure \ref{fig:numofcall} illustrates the performance versus the number of tokens used through OpenAI API, comparing \textsc{EvoKD} with AugGPT. The dataset is TouTiao news classification, under 1-shot.

It can be concluded that from the trending, our method has more advantages from a long-term perspective. In a single interaction, our method will consume more tokens than AugGPT because analyzing weaknesses and task explanations both require a significant number of tokens. However, as the number of interactions increases, the benefits brought by our method far outweigh the disadvantage of token consumption. After 10,000 tokens of interaction, the growth of AugGPT becomes very slow, while \textsc{EvoKD} is still growing.
In conclusion, AugGPT has a little superiority over \textsc{EvoKD} at the begining, while \textsc{EvoKD} ties AugGPT then and significantly outperforms AugGPT after about 8000-tokens interaction.

\subsection{Adaptability Analysis}

\begin{figure}[thb]
    \centering
    \includegraphics[width=\linewidth]{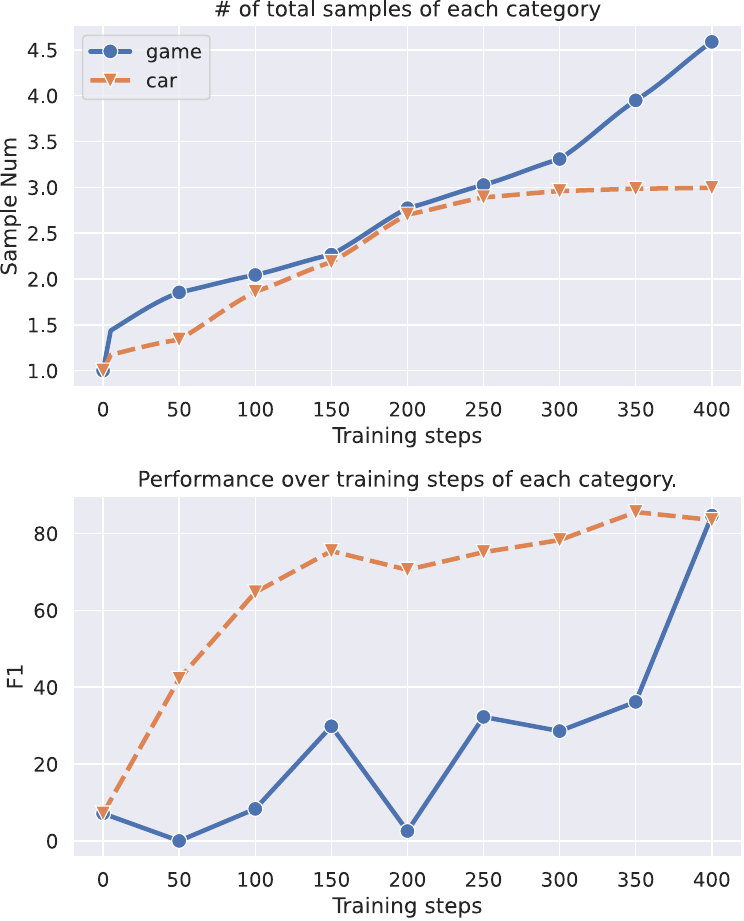}
    \caption{EvoKD concentrates on the samples with lower performance. Note that the upper sub-figure shows the accumulate number of samples of each category. A rising trending means the LLM generates more samples of the category, while a stage of horizontal line indicates that the category is absent in the generated data.}
    \label{fig:numeachcategory}
\end{figure}

We believe that the outstanding performance may come from the adaptability of \textsc{EvoKD}. In different environments, \textsc{EvoKD} should be able to adapt and adjust the strategy. To verify this, we take the category as the indicator, and study the performance changes of the student model on different categories during the training process.
% To investigate the reason why \textsc{EvoKD} has the outstanding performance, 
We plot the number of samples and F1 of each category on TouTiao news classification under 1-shot in Figure \ref{fig:numeachcategory}. We smooth the line of total number of samples. For better visualization and effectively highlighting the trends, two typical categories are chosen for comparison.%To clearly show the trending and distinguish the lines, 2 typical categories are selected to compare.

It is obvious that the performance on category ``game'' has lower F1 at the beginning, leading to an increase in the number of samples. With the increasing number of samples for ``game'', its F1 score eventually reaches approximately 80\% after 400 steps. In contrast, ``car'' maintains a consistently high level of performance, especially after 150 steps, resulting in a relatively stable number of instances. The increasing number of samples indicates that the LLMs concentrates on improving the performance on this category at that time point, thus generating more samples to teach the student. The active analysis can contribute to the F1 improvement of ``game''.

%accordingly, the number of samples increases. It can be seen that as the number of samples of ``game'' increasing, its F1 finally achieves about 80\% at 400 steps. On the contrast, ``car'' stays at a high level especially after 150 steps, so the number of instances is relatively stable. An increasing number of samples indicates that the LLMs concentrates on improving the performance on this category currently thus generates more samples to teach the student. The active analysis contributes to the improvement of ``game''.

\subsection{Few-Shot Classification}\label{sec:exp fewshot}

We conducted 3-shot and 5-shot text classification experiments and the results are shown in Table \ref{tab:few-shot cls exp}.

% \jiang{\sout{Obviously, ChatGPT augmentation substantially enhances the diversity of the training data.}}
In comparison to AugGPT, \textsc{EvoKD} exhibits higher F1 results and lower deviations across most of the datasets. In the case of TouTiao, \textsc{EvoKD} achieves respective scores of 0.7328 and 0.7474 under the 3-shot and 5-shot settings, which are higher than AugGPT by 3.4 percents and 2.3 percents.
\textsc{EvoKD} with initialization exhibits unstable superiority over bare \textsc{EvoKD}. For instance, \textsc{EvoKD} with initialization outperforms \textsc{EvoKD} by 4.5 percents on Amazon, but underperforms it by 7.6 percents on IMDB.
% Additionally, we notice that the advantage of our proposed method becomes less as the number of shots increases. This phenomenon could be expected, since an increase in the number of samples leads to a larger pool of data available for text augmentation, and the quality of the augmented data is not substantially affected by using different DA approaches.

\subsection{1-Shot NER}\label{sec:exp gen}

\begin{table}[thb]
    \centering
    \footnotesize
    \begin{tabular}{m{1.7cm}m{1.4cm}<{\centering}m{1.4cm}<{\centering}m{1cm}<{\centering}}
    \toprule
    Method & CoNLL03 & CoNLL04 & AVG\\
    \midrule
        Full Shot & 0.9322 & 0.8766 & 0.9044 \\
    \midrule
        No Augment & 0.3143 &	0.4929 & 0.4036 \\
        EDA & 0.3062 & 0.5058 & 0.4060\\
        AugGPT & 0.6315 & \underline{0.6683} & 0.6499\\
        \midrule
        \textsc{EvoKD} & \underline{0.6538} & \textbf{0.6848} & \textbf{0.6693}\\
        \quad +Init & \textbf{0.6629} & 0.6628 & \underline{0.6629}\\
    \bottomrule
    \end{tabular}
    \caption{1-shot results for NER datasets, where AVG denotes the average F1 performance over CoNLL03 and CoNLL04.}
    \label{tab:ner}
\end{table}

% \begin{table}[htb]
%     \centering
%     \footnotesize
%     \begin{tabular}{m{1.7cm}m{1.4cm}<{\centering}m{1.4cm}<{\centering}m{1cm}<{\centering}}
%     \toprule
%     Method & CoNLL03 & CoNLL04 & AVG\\
%     \midrule
%         Full Shot & 0.9322 & 0.8766 & 0.9044 \\
%     \midrule
%         No Augment & 0.3143 $\pm$0.1119 &	0.4929 $\pm$0.0705 & 0.4036 \\
%         EDA & 0.3062 $\pm$0.1330	& 0.5058 $\pm$0.0784 & 0.4060\\
%         AugGPT & 0.6315 $\pm$0.0593 & \underline{0.6683} \underline{$\pm$0.0097} & 0.6499\\
%         \midrule
%         \textsc{EvoKD} & \underline{0.6538} \underline{$\pm$0.0233}	& \textbf{0.6848 $\pm$0.0150} & \textbf{0.6693}\\
%         \quad +Init & \textbf{0.6629 $\pm$0.0289} & 0.6628 $\pm$0.0184 & \underline{0.6629}\\
%     \bottomrule
%     \end{tabular}
%     \caption{1-shot results for NER datasets, where AVG denotes the average F1 performance over CoNLL03 and CoNLL04.}
%     \label{tab:ner}
% \end{table}

% We investigate the performance of \textsc{EvoKD} on NER datasets in this SubSection.
% The entity types "other" and "miscellaneous" were excluded because they do not have clear and specific meanings. 
We consider a prediction to be correct only if both the entity type and the entity text align with the ground truth. Based on the metric, we report the F1 results in Table \ref{tab:ner}.

Generally, the 1-shot performances without KD or EDA are unsatisfactory. Leveraging the generation ability, AugGPT significantly enhances the average result to 0.65, outperforming EDA by a considerable margin.
\textsc{EvoKD} achieves the best perfomance. It outperforms AugGPT on CoNLL04 by approximately 2 percents, while \textsc{EvoKD} with AugGPT initialization achieves the highest F1 on CoNLL03, surpassing AugGPT by around 3 percents.%\textsc{EvoKD} outperforms bare ChatGPT augmentation on CoNLL04 by approximately 2 percents, while \textsc{EvoKD} with ChatGPT initialization achieves the highest F1 on CoNLL03, surpassing bare ChatGPT by around 3 percents.

% We notice that comparing with text classification, \textsc{EvoKD} has only slight improvement on NER datasets. This is because that 1) constructing NER samples is more challenging than classification, and large language model may be relatively inadequate in Information Extraction tasks. 2) Furthermore, through analyzing the augmented samples, we have discovered that the performance of NER is heavily dependent on the knowledge distribution within the dataset, and the augmented texts may not be able to encompass all the knowledge tuples. For instance, the augmented text may include the country ``Japan'', while the base model is tested with the country ``Uzbekistan'', making it challenging to extract the correct entity type and span.

\subsection{Ablation Study}\label{sec:exp pipeline}

\begin{table}[thb]
    \centering
    \footnotesize
    \begin{tabular}{lm{1cm}<{\centering}m{1cm}<{\centering}m{1cm}<{\centering}}
        \toprule
        Method & Amazon & Inshorts & AVG\\
        \midrule
        % \textsc{EvoKD} & \textbf{0.8425 $\pm$ 0.0317} & \textbf{0.8516 $\pm$ 0.0257} & \textbf{0.8471} \\
        % \quad w/o Easy & 0.5976 $\pm$ 0.0952 & 0.6491 $\pm$ 0.1553 & 0.6234\\
        % \quad w/o Correct & 0.8132 $\pm$ 0.0373 & 0.8334 $\pm$ 0.0247 & 0.8233\\
        % \quad w/o Review & 0.7324 $\pm$ 0.0948 & 0.8152 $\pm$ 0.0345 & 0.7738 \\
        % \quad w/o Separating Text \& Label & 0.7112 $\pm$ 0.1336 & 0.8359 $\pm$ 0.0367 & 0.7736 \\
        \textsc{EvoKD} & \textbf{0.8425} & \textbf{0.8516} & \textbf{0.8471} \\
        w/o Easy & 0.5976 & 0.6491 & 0.6234\\
        w/o Correct & 0.8132 & 0.8334 & 0.8233\\
        w/o Review & 0.7324 & 0.8152 & 0.7738 \\
        w/o Separating & 0.7112 & 0.8359 & 0.7736 \\
        \bottomrule
    \end{tabular}
    \caption{Ablation Study on 1-shot text classification. We remove $D_{easy}$ and $D_{correct}$ respectively, denoted as ``w/o Easy'' and ``w/o Correct'' respectively. In addition, we drop the strategy of ``Review History'', which is denoted as ``w/o Review''. And we also merge sub-steps ``Input Text Generation'' and ``Labeling'' discussed in SubSection \ref{subsec:chat pipeline} together, which is denoted as ``w/o Separating''.}
    \label{tab:ablation}
\end{table}

We conduct ablation study on our pipeline and the prompt, the results are shown in Table \ref{tab:ablation}.
We find that both easy samples and correct samples enhance the quality of the generated texts. 
Overall, the former is more crucial. The risk associated with removing the easy samples generated by the LLM is more substantial, resulting in a decrease of 0.2237 on average F1.
%The former is more crucial overall. The harm of removing easy samples generated by ChatGPT is conceivable, with a decrease of 0.2237 on the average F1.
% \sihong{better to explain from the aspect of catastrophic forgetting and distribution of hard and easy samples. sentence below is kind of difficult to understand. }
As mentioned in Section \ref{sec:actionda}, training only on $D_{hard}$ would cause the problem of catastrophic forgetting. The student model would learn the biased distribution.
% ChatGPT may guess wrong weaknesses of the base model and focus on generating actually easy or not very hard cases. Consequently, the augmented texts would not provide much compensation for the base model's deficiencies.

Across the pipeline, we remove 1) the strategy ``Review History'' and 2) merge the sub-steps ``“Input Text Generation'' and ``Labeling''. Their performance degradation are roughly equal, about 7.3 percents, which indicates that they are equally important.

\section{Conclusion}

In this paper, we propose \textsc{EvoKD}: Evolving Knowledge Distillation with LLM and Active Learning, which is an effective framework especially for few-shot setting. We prompt ChatGPT to analyse the weakness of the student model, and subsequently generate samples based on the analysis. The experimental results demonstrate the effectiveness of \textsc{EvoKD}, particularly on most text classification datasets where we achieve 90\% of the full-shot performance with only 1-shot.
% \textsc{EvoKD} also has lower deviation using different seed samples. Our objective is to explore to what extent the abilities of generation, reasoning, and knowledge of LLMs enhance DA, rather than simply utilizing LLMs as text rewrites or arbitrary text generators. We hope that our work will stimulate academic attention to the use of LLMs for \naishan{few-shot} data augmentation.

\section{Limitations}

% From the experiments we found that, \textsc{EvoKD} suffer from heavy cost of web communication, like other approaches using LLMs. Most of the training time was spent on waiting for the response from ChatGPT, causing a low GPU utilization. Although our proposed strategies alleviate the time-consuming interaction, the whole process still cost much longer time than normal few-shot training without ChatGPT. Future works may explore fast text augmentation with lower time cost.

We notice that the advantage of our proposed method becomes less as the number of shots increases. This phenomenon could be expected, since an increase in the number of samples leads to a larger pool of data available, and the quality of the generated data is not substantially affected by different KD approaches.

% Besides, the experimental results show that \textsc{EvoKD} achieves only limited improvement on the task of NER. After analyzing the results, we find that ChatGPT often tends to summarize imprecise weakness of the base NER model, which leads to incomplete coverage of entities and world knowledge in the enhanced text.

\section{Ethics Statement}
We introduce \textsc{EvoKD} in this paper. Our design of the prompt texts and samples generated by ChatGPT honors the ethical code. In our experiment, ChatGPT is applied, which is a large language model pre-trained on a large-scale corpus. We encourage researchers to explore whether ChatGPT-generated data is biased and discriminatory before deploying \textsc{EvoKD} to ethically improve the performance of student models.

\section{Acknowledgements}
This work was supported in part by National Key Research and Development Program of China (2022YFC3340900), National Natural Science Foundation of China (62376243, 62037001, U20A20387), the StarryNight Science Fund of Zhejiang University Shanghai Institute for Advanced Study (SN-ZJU-SIAS-0010), Alibaba Group through Alibaba Research Intern Program, Project by Shanghai AI Laboratory (P22KS00111), Program of Zhejiang Province Science and Technology (2022C01044).

% \nocite{*}
% \section{Bibliographical References}\label{sec:reference}

% \bibliographystyle{lrec-coling2024-natbib}
% \bibliography{lrec-coling2024-example}

% \section{Language Resource References}
% \label{lr:ref}
% \bibliographystylelanguageresource{lrec-coling2024-natbib}
% \bibliographylanguageresource{languageresource}

\nocite{*}
\section{Bibliographical References}\label{sec:reference}

\bibliographystyle{lrec-coling2024-natbib}
\bibliography{lrec-coling2024-example}

\section{Language Resource References}
\label{lr:ref}
\bibliographystylelanguageresource{lrec-coling2024-natbib}
\bibliographylanguageresource{languageresource}

\appendix

\section{Interaction Example}
\label{sec:chat example}

\begin{figure*}[thb]
    \centering
    \includegraphics[width=\linewidth]{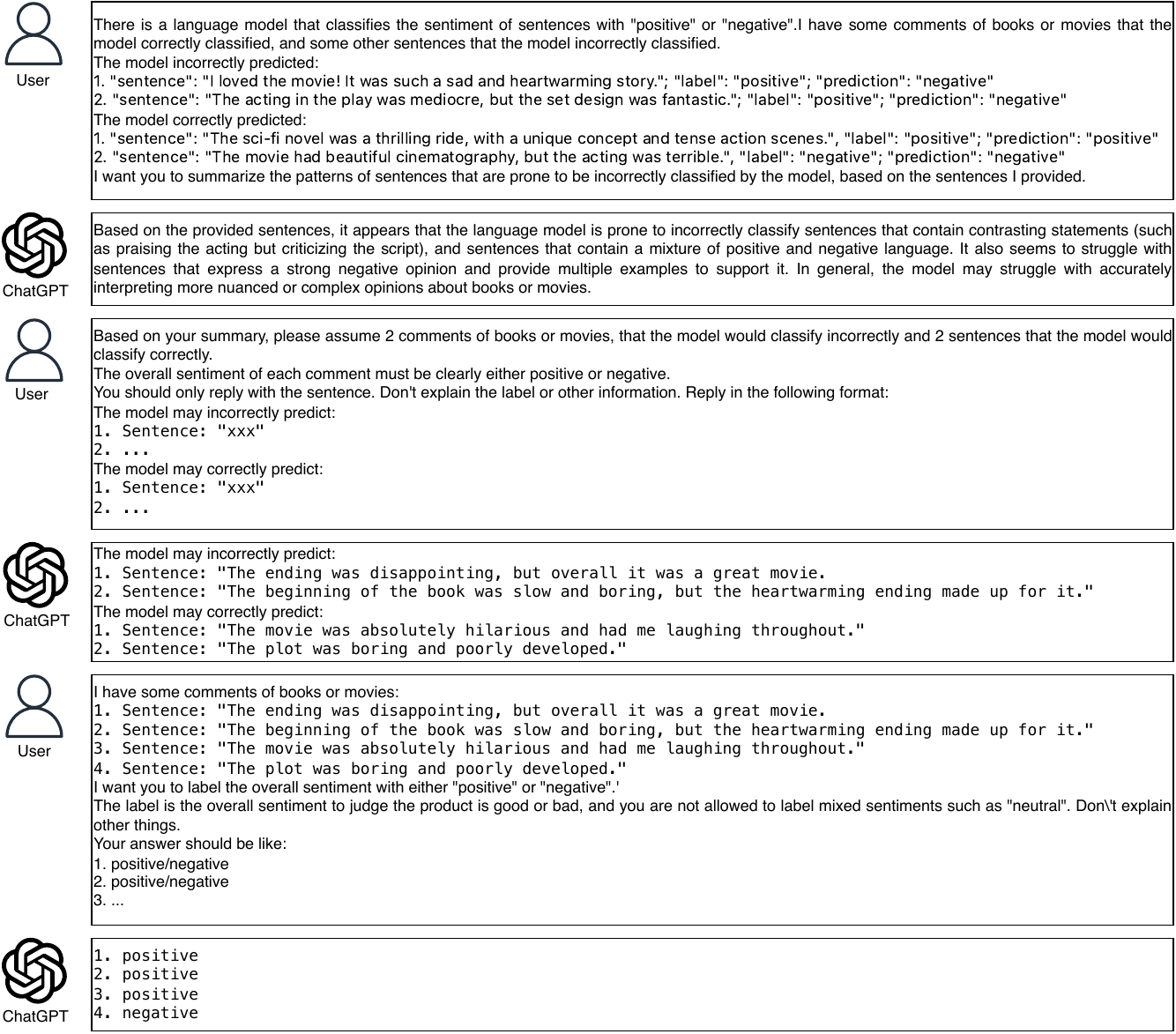}
    \caption{Example of evolving active learning of sentiment classification task.}
    \label{fig:chat}
\end{figure*}

We take the sentiment classification as an example and demonstrate our Evolving Active Learning pipeline in Figure \ref{fig:chat}.

\section{ChatGPT Augmentation Example}
\label{sec:ChatGPT Augmentation Example}

We list some examples by bare ChatGPT augmentation of sentiment classification in Table \ref{tab: chatgpt aug example}.

\begin{table*}
    \centering
    \begin{tabular}{p{8cm}|p{8cm}}
    \toprule
        \textbf{Original Text} & \textbf{Augmented Text} \\
    \midrule
        I purchased this item after reading reviews. Using the directions for roasting in the microwave the garlic was burnt to a crisp, the smell lingered for days. I immediately threw this garlic roaster in the trash. I will use foil and bake in the oven. What a waste of money. & The hype around this restaurant was unwarranted - the food was overpriced and underwhelming. I regret spending my money here.\\
        \midrule
        Kylie is back in the States! Though she was never missing from the pop music world always making music... I like it, and I think it is some of the best dance music to come from Kylie in a while. & Kylie's latest single ``Fever'' is a subtle but catchy dance song with rhythmic beats and a hook that grows on you - some of the best dance music from her in a while, and I like it!\\
        \midrule
        i ordered five of these for my grandkids because I saw an ``8 x 10.5'' book and the contents are great- it looked wonderful! BUT, when they arrived I was very disappointed to see... The website does not show the size of the book. Who makes childrens books in what must be about a size 6 font? I feel like I wasted my money.For heavens sake put the size on your books! & The product photos looked fantastic, but the actual item was a disappointment, as it was much smaller than expected and its features became hard to see.\\
        \midrule
        I was held completely enthralled through the first half of this book. i loved the setting and it's description. I loved the heroine and the ``wee lad'' she took under her wing. My only problems with it were that, for me, there weren't enough love scenes, and they arrived too late in the story...and the fact the author decided to throw in another couple into the mix, without fully fleshing out their story. I think Ms. Garwood should have stuck to her main protagonists and left the other couple for another book & I thoroughly enjoyed the storyline and well-rounded characters, but would have preferred a stronger focus on the central romance and less on the supporting characters.\\
    \bottomrule
    \end{tabular}
    \caption{Examples augmented by bare ChatGPT.}
    \label{tab: chatgpt aug example}
\end{table*}

\end{document}